%% file: main.tex
\def\year{2021}\relax
\documentclass[letterpaper]{article} 
\usepackage{aaai21}  
\usepackage{times}  
\usepackage{helvet} 
\usepackage{courier}  
\usepackage[hyphens]{url}  
\usepackage{graphicx} 
\usepackage{natbib}  
\usepackage{caption} 
\usepackage{epstopdf}
\usepackage{xcolor}

\usepackage{amsfonts}
\usepackage{amsmath}
\usepackage{booktabs}
\usepackage{subcaption}
\usepackage{placeins}

\title{Predicting Livelihood Indicators from Community-Generated Street-Level Imagery}
\author{

    Jihyeon Lee, \textsuperscript{\rm 1}
    Dylan Grosz, \textsuperscript{\rm 1}
    Burak Uzkent, \textsuperscript{\rm 1}
    Sicheng Zeng, \textsuperscript{\rm 1}\\ 
    Marshall Burke, \textsuperscript{\rm 2}
    David Lobell, \textsuperscript{\rm 2}
    Stefano Ermon \textsuperscript{\rm 1}
    \\
}
\affiliations{

    \textsuperscript{\rm 1}Department of Computer Science, Stanford University\\
    \textsuperscript{\rm 2}Department of Earth Science, Stanford University\\

    \{jihyeon,dgrosz,buzkent\}@cs.stanford.edu\\

}

\begin{document}

\maketitle

\begin{abstract}
Major decisions from governments and other large organizations rely on 
measurements of the populace's well-being, 
but making such measurements at a broad scale is expensive and thus infrequent in much of the developing world. We propose an inexpensive, scalable, and interpretable approach to predict key livelihood indicators from public crowd-sourced street-level imagery. 
Such imagery can be cheaply collected and more frequently updated compared to traditional surveying methods, while containing plausibly relevant information for a range of livelihood indicators. 
We propose two approaches to learn from the street-level imagery: (1) a method that creates multi-household cluster representations by detecting informative objects and (2) a graph-based approach that captures the relationships between images.
By visualizing what features are important to a model and how they are used, we can help end-user organizations understand the models and offer an alternate approach for index estimation that uses cheaply obtained roadway features.
By comparing our results against ground data collected in nationally-representative household surveys, we demonstrate the performance of our approach in accurately predicting indicators of poverty, population, and health and its scalability by testing in two different countries, India and Kenya. Our code is available at  https://github.com/sustainlab-group/mapillarygcn.
\end{abstract}

\section{Introduction} \label{intro}
\input{sections/intro.tex}
\section{Related Works} \label{relatedworks}
\input{sections/related_works}

\section{Datasets} \label{dataset}
\input{sections/dataset.tex}

\section{Methods}
\input{sections/methods.tex}

\section{Experiments} \label{experiments}
\input{sections/experiments.tex}

\begin{figure*}[!t]
\centering
\includegraphics[width=0.93\linewidth]{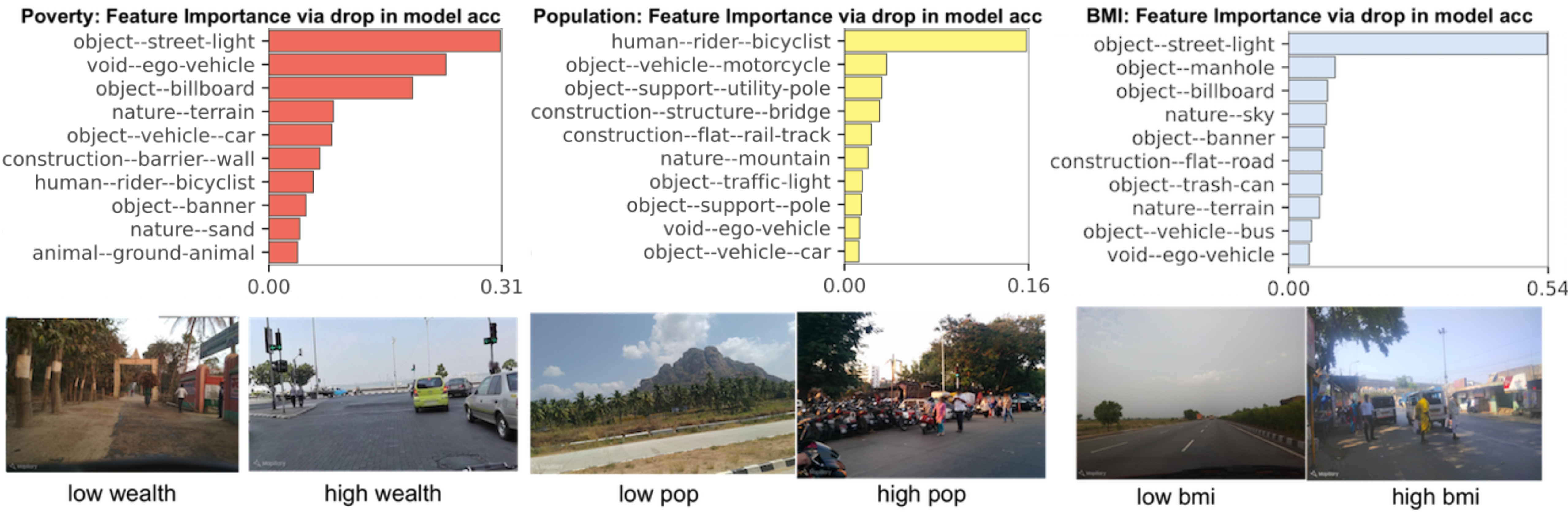}
\caption{Top: Feature importance by indicator. We train a pooled random forest model, using data from both countries, for poverty and population. Women's BMI data was not available in Kenya, so the BMI features are India-specific. We use permutation importance because it is more reliable than mean-decrease-in-impurity importance \protect\cite{Strobl2007}, which can be highly biased when predictors variables vary in their scale of measurement or number of categories, as in our case.}
 \label{fig:mostimportantfeats}
\end{figure*}

\begin{figure*}[!h]
\centering
\includegraphics[width=0.9\linewidth]{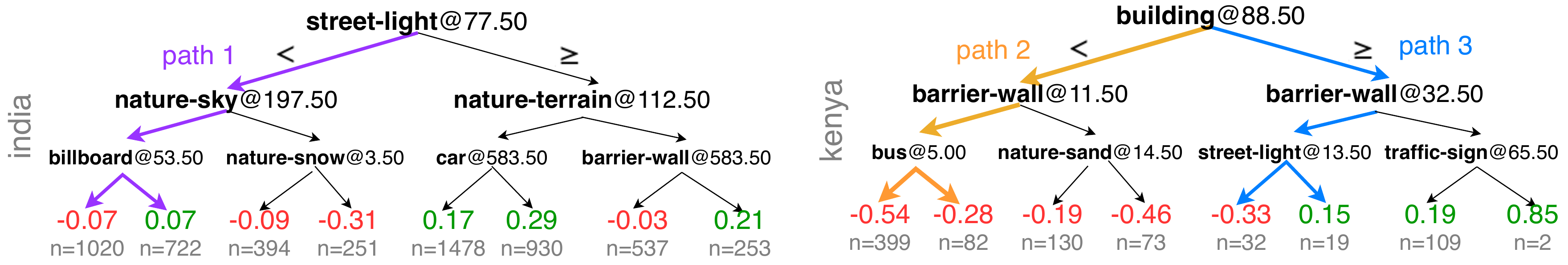}
 \caption{Decision tree visualizations. A decision tree classifier is trained for poverty regression in each country and then visualized \footnote{https://github.com/parrt/dtreeviz}. Each node displays the object class and threshold that determines how to split the node (left child means $<$ the threshold, right child $\geq$ the threshold). Visualizations for all indicators in Supplementary Material.
 }
 \label{fig:trees}
\end{figure*}

\section{Interpretability using Object Counts} \label{sec:interpretability}
\input{sections/interpretability.tex}

\section{Conclusion}
\input{sections/conclusion.tex}


\section{Acknowledgments}
This research was supported by the Stanford King Center on Global Development and NSF grants \#1651565 and \#1733686.

\bibliography{references}

\end{document}

%% file: sections/intro.tex
In 2015, all member states of the United Nations adopted 17 Sustainable Development Goals \footnote{https://sdgs.un.org}, including eliminating poverty, achieving good health, and stimulating economic growth by 2030. To evaluate countries' progress toward these goals, national governments and international organizations conduct nationally-representative household surveys that collect information on a range of livelihood indicators from households distributed throughout a given country. These surveys, such as the Demographic and Health Surveys (DHS) Program, provide critical insight into local economic and health conditions \footnote{\url{https://dhsprogram.com/What-We-Do/Survey-Types}}. However, they are costly and time-consuming, particularly when surveying remote populations linked by poor infrastructure.  
As a result, surveys are conducted infrequently or may only capture an extremely small proportion of households~\cite{yeh2020using}. Satellite images and machine learning have been proposed as an alternative~\cite{ayush2020generating,ayush2020efficient,uzkent2020efficient,uzkent2019learning,Jean790}, but high-resolution imagery is expensive and obscures details of the ground level.


We introduce a scalable, interpretable approach that uses street-level imagery for livelihood prediction at a local level.
We utilize Mapillary, a global citizen-driven street-level imagery database~\cite{MVD2017}. 
Although Mapillary cannot match the consistent quality of commercial imagery, its widespread and growing availability in developing regions make it an appealing candidate as a passively collected data source for predicting livelihood indicators. In eight months, the Mapillary dataset doubled in size from 500 million images to 1 billion, with users capturing and verifying imagery from mobile devices \footnote{\url{https://help.mapillary.com/hc/en-us/articles/115001478065-Equipment-for-capturing-and-example-imagery}}.

We show how to capture information from Mapillary imagery to accurately predict livelihood indicators in India and Kenya, some of the most populous and economically diverse countries in the world. We present two complementary approaches: 
(1) The first creates representations for multi-household clusters by segmenting street-level images and aggregating informative objects, trains models, and interprets them using the most predictive features. 
(2) While the strength of the first approach is interpretability, we also propose a second to learn the relationships between images and leverage the inherent spatial structure of clusters by representing them as graphs, where each image is a feature-rich node connected by edges based on spatial distance.

Our approaches predict three indicators --- poverty, population, and women's body mass index (BMI, a key nutritional indicator) --- in villages and urban neighborhoods. They achieve high classification accuracy and strong $r^2$ scores for regression in India and Kenya. 
Our method is a cheap, scalable, and effective alternative to traditional surveying to measure the well-being of developing regions. 

%% file: sections/related_works.tex
Recent research has explored the usage of passively collected data sources as cheaper alternatives to door-to-door or paper forms of data collection, to augment or eventually replace expensive household surveys. Proposed sources include social media~\cite{signorini2011use,pulse2014mining}, mobile phone networks~\cite{blumenstock2015predicting}, Wikipedia~\cite{sheehan2019predicting}, remote sensing, and street-level images. 
\subsection{Remote Sensing Data} Remote-sensing imagery from satellite or aircraft has been used to predict road quality in Kenya~\cite{cadamuro2019street}, land use patterns in European cities~\cite{Albert2017LandUse}, and economic outcomes in Africa and India~\cite{Jean790,yeh2020using,Pandey2018,ayush2020efficient}. However, these approaches face challenges, such as poor generalizability to other locations or indicators~\cite{head2017can,bruederle2018nighttime} and lack of nuance as local scenes are not visible from space. Street-level imagery provides greater detail (e.g. people) and local information.
\subsection{Street-level Imagery} Past studies have used street-level imagery to measure social or health outcomes, but the focus has been in urban areas, predicting perceived safety of American cities with Google StreetView~\cite{Fu2018Streetnet, naik2014streetscore}, analyzing urban perception from Baidu Maps or Tencent Maps in Chinese cities~\cite{Gong2019Beijing,yao2019urbanperception}, identifying urban areas from human ratings of StreetView images in India~\cite{galdo2020urbanindia}, or predicting crime~\cite{crimerateChicago} and voting patterns~\cite{Gebru13108} in U.S. cities. Our goal is to create approaches for \textit{developing} regions that lack infrastructure by leveraging a community-generated data source, with an additional challenge that crowdsourced data can be noisy and inconsistent.


Another study that utilizes street-level imagery is \cite{Suel2019london}, which trains a CNN to predict the best and worst-off deciles for outcomes like crime and self-reported health (it performs poorly for predicting objective health). The authors trained the CNN on Google StreetView imagery from London and demonstrated that learning could transfer to three other cities of the UK~\cite{Suel2019london}. We try the same method in our experiments, training a CNN on Mapillary images. However, given the noisy quality of the dataset and more difficult task of predicting accurately throughout developing countries and not just within developed cities, we require more sophisticated approaches to make improvements. For example, our graph convolutional network learns from multiple images, representing the spatial structure between them in edges, and encodes useful semantic features in the nodes. Furthermore, the visual makeup of cities in the same country can be similar, and it is a harder task for models to generalize to other countries. By performing experiments in India and Kenya, which have different data distributions, we show our approach can scale across countries. 

Lastly, there are works that use crowdsourced imagery, i.e. Flickr, to predict the ambiance of London neighborhoods \cite{redi2018ambiance} and "geo-attributes" such as GDP for gridded cells across the world \cite{lee2015predicting}. The latter is the most similar work since authors use a crowdsourced dataset to make predictions at a global, not solely urban, level. However, only 5\% of Flickr's imagery is geotagged~\cite{hauff2013}. We  found Mapillary is a more apt dataset for our work given its coverage of developing areas, exponential growth~\cite{solem_2019}, and geotagged imagery. 



%% file: sections/dataset.tex
First, we define the general problem of making predictions on geospatially located clusters. Assume there are $N$ geographic clusters. A cluster $i$ represents a circular area with center $c_{i}=(lat_{i}, lon_{i})$. There is a set of street-level images that fall within its spatial boundaries, $\mathcal{X}_i = \{x_i^{0},x_i^{1},...,x_i^{n_i}\}$, where each $x_i^{j} \in \mathcal{X}$ is an image and $n_i$ is the total number of images. Each cluster also has $K$ surveyed variables of livelihood indicators, represented by $y_i$. 
\begin{figure*}[!t]
\centering
\begin{tabular}{ccc} 
\includegraphics[width=0.88\textwidth]{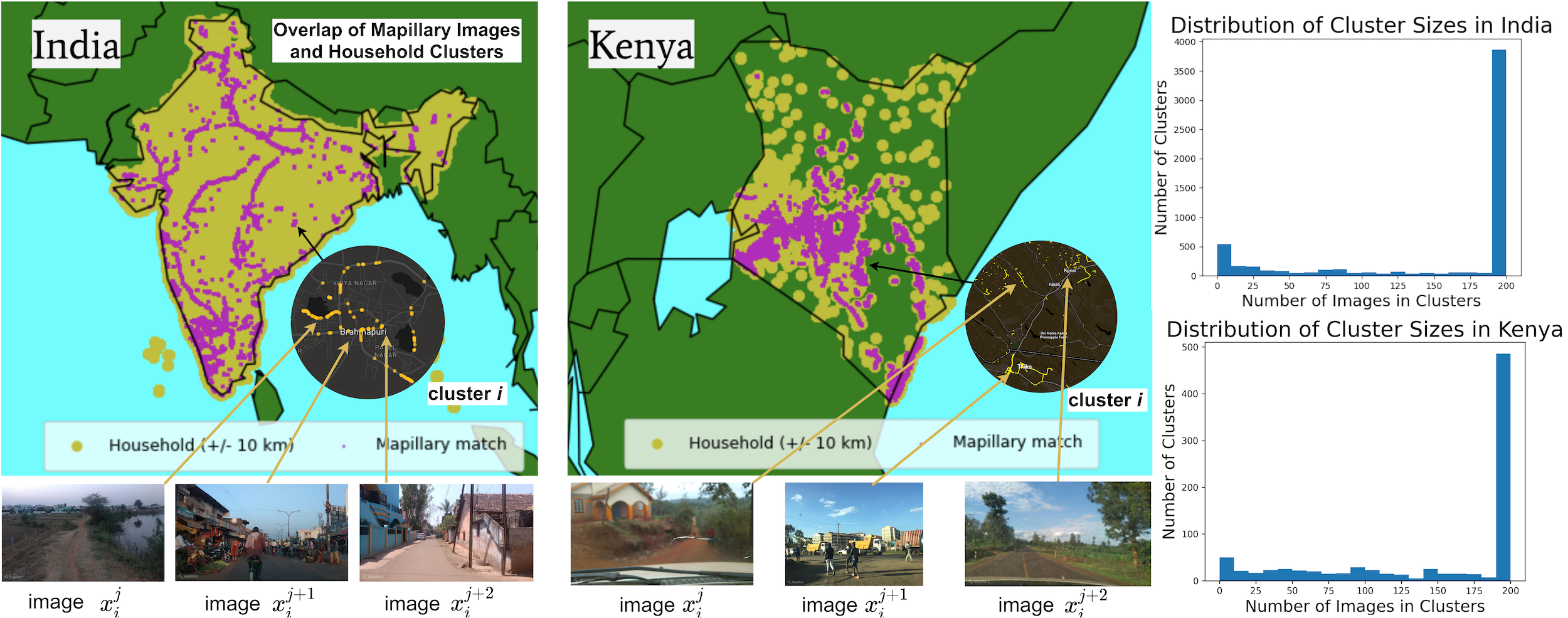}
\end{tabular}
\caption{Left: Overlap of Mapillary images with household clusters for India and Kenya. Each map also contains a close-up view of a cluster $i$, where yellow dots represent its images. Right: Distribution of number of images per cluster in either country.}
\label{fig:coverage}
\end{figure*}
\begin{figure*}[!t]
\centering
\begin{tabular}{ccc} 
\includegraphics[width=0.95\textwidth]{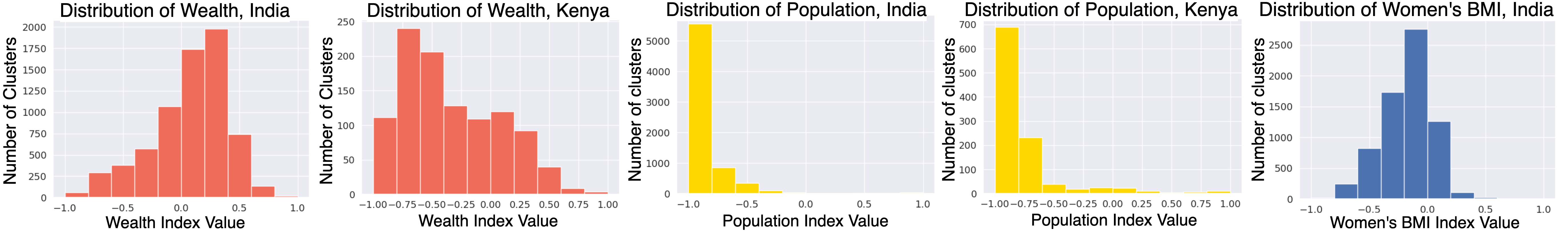}
\end{tabular}
\caption{Left to right: Distributions of wealth (poverty is defined as its inverse) and population over India and Kenya. Distribution of women's BMI in India. Note that binary labels are generated from a median split, so classes are balanced.}
\label{fig:distribution}
\end{figure*}
We aim to learn a mapping: $\mathcal{P}(\mathcal{X}) \mapsto \mathcal{Y}$ to predict $y_{i}$ given $\mathcal{X}_i$, where $\mathcal{P}(\mathcal{X})$ is the powerset of $\mathcal{X}$ and $\mathcal{X}_i \in \mathcal{P}(\mathcal{X})$. The regression task entails predicting the indicator directly. 
Classification involves predicting a binary label, where an indicator value $\geq$ to the median results in a label of 1 and 0 otherwise. We perform these tasks over image sets $\mathcal{X}_i \in \mathcal{P}(\mathcal{X})$ as clusters have a variable number of images.

We specialize the general problem to a dataset of street-level imagery and cluster-level labels of indicators in India and Kenya. Each cluster represents a group of households within a 5km-radius area with label $y_i$, which contains the index value and class label for $K$ indicators. In India, $K=3$, i.e. poverty, population, and women's BMI. In Kenya, $K=2$ as women's BMI was not available. We discuss the data sources of $\mathcal{X}$ and $\mathcal{Y}$ in the following section. 

\subsection{Mapillary for Street-Level Images}
The Mapillary API \footnote{\url{https://www.mapillary.com/developer}} provides access to geo-tagged images and metadata. The data is community-driven, and anyone can create an account and upload images with EXIF embedded GPS coordinates or identify the location on a map. We matched images to household clusters, and only clusters with $n_i > 0$ were kept. This resulted in 7,117 clusters and 1,121,444 images from India and 1,071 clusters and 156,756 images from Kenya. 98\% of Mapillary images were available in high-resolution ($2048 \times 1536$ px) and the remaining 2\% in lower resolution ($640 \times 420$ px). 
\subsection{Livelihood Indicators} \label{dhs}
We predict three varied livelihood indicators: poverty and population in both countries and a health-related measure in India. Each index is naturally continuous. The values are rescaled to be between -1 and 1 and used directly for regression. We split by the median to produce class labels. Figure \ref{fig:coverage} shows the overlap of Mapillary images and indicator labels, and Figure \ref{fig:distribution} shows the index distribution by country. 
\subsubsection{Poverty} We obtained wealth index values from the most recently completed surveys of the Demographic and Health Survey (DHS), 2015-16 for India and 2014 for Kenya. DHS data is clustered; households within a 5km-radius contribute datapoints individually but share the same geographic coordinates to preserve privacy.
The index is calculated from assets and characteristics, such as vehicles, home construction material, etc. We consider poverty as the inverse of wealth. 
\subsubsection{Population} Facebook's High Resolution Population Density Maps consist of geo-located population density labels across the world. Their data is much denser than that of Mapillary Vistas, so we average the values within a 5km radius of a cluster's coordinate to produce its label. 
\subsubsection{Women's BMI} Women's body-mass-index (BMI) is an important indicator of human well-being. We compute BMI by dividing weight in kilograms by height in meters squared for the 697,486 samples in the DHS survey and average the values across all women in a cluster to produce its label. 



%% file: sections/methods.tex
Given 
this dataset 
constructed with geotagged Mapillary images and indexes, we propose methods to learn the mapping  $\mathcal{P}(\mathcal{X}) \mapsto \mathcal{Y}$. We focus on two learning paradigms: (1) image-wise learning where a model learns from a single image, $x_{j} \in \mathcal{X}_{i}$, sampled from cluster $c_{i}$, and (2) cluster-wise learning where a model learns from all images $\mathcal{X}_{i}$ in $c_{i}$.
\begin{figure*}[!ht]
\centering
\includegraphics[width=0.9\linewidth]{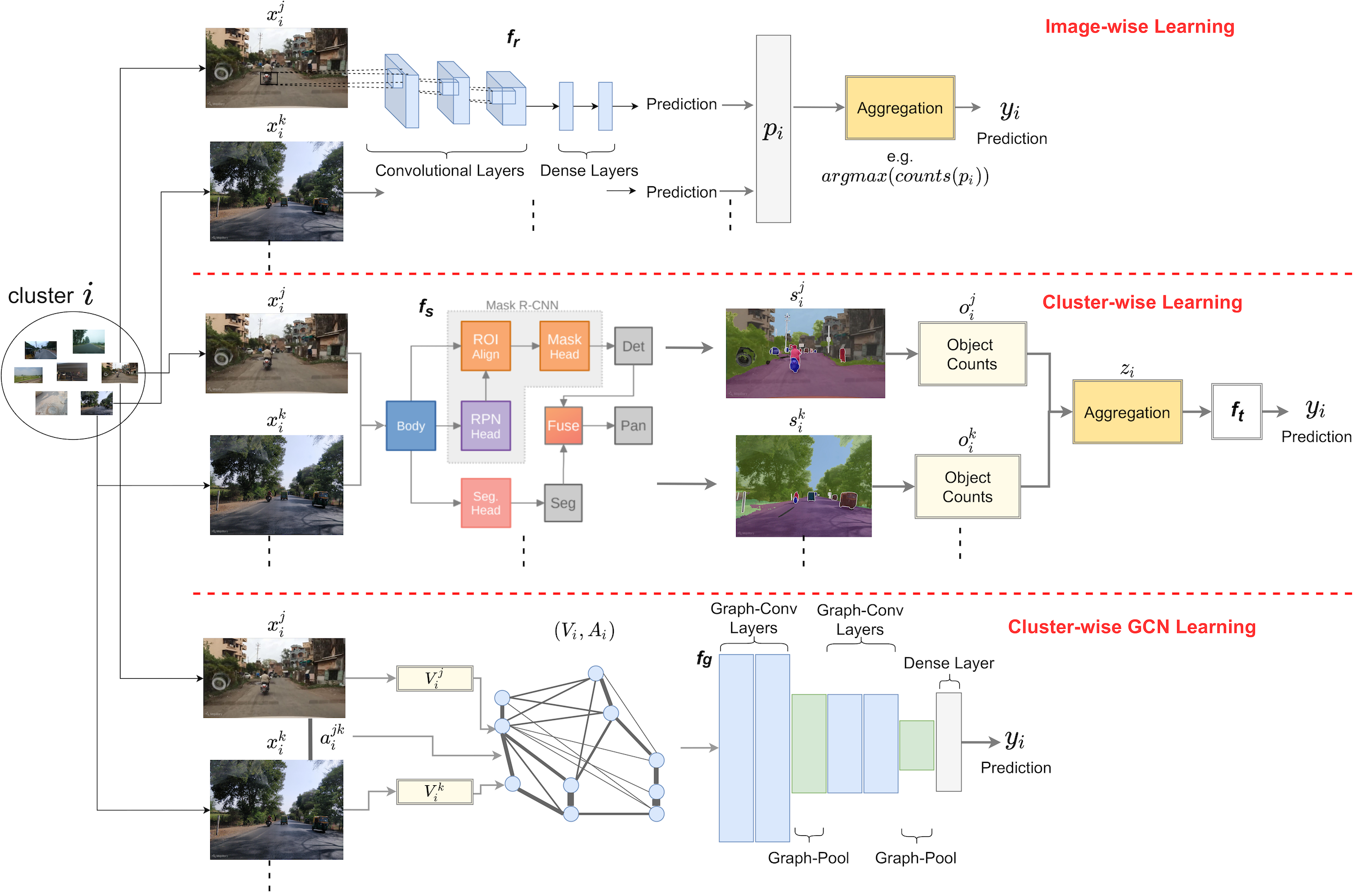}
\caption{Overview of the proposed methods. Top: image-wise learning, which learns mapping $f_r$ from imagery as in \cite{Suel2019london}. Middle: cluster-wise learning, which uses a panoptic segmentation model to create cluster-level representations and learns the mapping $f_t$. Bottom: GCN that represents clusters as a fully connected set of images. Some connections are left out in the above figure for the sake of clarity. Using Graph-Conv layers followed by a Graph-Pool layer, it learns mapping $f_g$.}
\label{fig:pipeline}
\end{figure*}
\subsection{Image-wise Learning}
As did \cite{Suel2019london}, we directly map each individual image in the cluster, $x_{j} \in \mathcal{X}_i$, to the label space.
We refer to the model that learns this mapping as \textbf{ResNet-ImageWise}. As in Figure \ref{fig:pipeline}, image-wise predictions, $p_i$, are combined at test time using an aggregation strategy to produce final predictions, $y_i$. 
Each prediction is considered a vote, and the majority class is considered the final prediction for cluster $i$.
\subsection{Cluster-wise Learning}
\subsubsection{Learning from Cluster Level Object Counts}
In this section, we propose a method to utilize object counts from street-level images in a cluster. Image level object counts are aggregated across the cluster to create cluster-wise object counts. Finally, we train a classifier or regression model on cluster-level object counts to predict indexes.
\subsubsection{Panoptic Segmentation on Mapillary Images}
\label{sect:Panoptic_Segmentation}
With the Mapillary Vistas~\cite{MVD2017} panoptic segmentation dataset, we can train a network to segment street-level images. Mapillary Vistas has 28 \emph{stuff} and 37 $\emph{thing}$ classes, where \emph{stuff} refers to amorphous regions, i.e. "nature," and \emph{things} are enumerable objects, i.e. "car." It contains 25,000 annotated images with an average resolution of $\sim$9 megapixels captured at various conditions, times, and viewpoints (e.g. from a windshield, while walking, etc.), making Mapillary Vistas an ideal dataset to train a segmentation model. 

We use the seamless scene segmentation model proposed by~\cite{Porzi_2019_CVPR}. The model consists of two main modules--instance segmentation and semantic segmentation--and the third module fuses predictions from both to generate panoptic segmentation masks. 
The instance segmentation module uses Mask-RCNN~\cite{he2017mask}, and the semantic segmentation module uses an encoder-decoder architecture similar to the Feature Pyramid Network~\cite{lin2017feature}. Finally, ResNet50 is used to parameterize the backbone model. During training, the Mapillary images are scaled such that the shortest size is $1024 \times t$ pixels, where $t$ is randomly sampled from the interval $[0.5,2.0]$. 
The authors report $50.4\%$ mean IoU (intersection over union) score on the Mapillary Vistas test set. To be consistent with the trained model, we scale our Mapillary images such that the shortest size is represented by 1024 px. 
\subsubsection{Cluster Level Object Counts}
Using the seamless scene segmentation model, we segment every image $x_i^j \in X_i$ in a cluster with the hypothesis that the 65 different roadway features provide useful information. The authors of
\cite{ayush2020generating} correlated object detections from satellite imagery with poverty in Uganda and used them as interpretable features. We expected to discover patterns as well, such as more bike racks appearing in high-wealth areas. Each image $x_i^{j} \in \mathcal{X}_i$ maps to a set of object detections $o_i^{j}$, where $o_i^{j} \in \mathbb{R}^{65}$. We then sum the number of instances for each class, or $\sum_{j=0}^{n_i} o_i^{j}$. To avoid bias towards clusters with many images, we append a feature with the total number of images in a cluster, or $n_i$. 
Each cluster is represented by a feature vector $z_i \in \mathbb{R}^{66}$.
We map these interpretable embeddings to the label space and refer to the models as \textbf{Obj-ClusterWise}. 
\subsection{Graph Convolutional Networks}
The methods thus far process images independently. We propose Graph Convolutional Networks (GCN) \cite{DBLP:journals/corr/SuchSDPZMCP17} to learn relationships between images, representing a cluster as a graph, where image-based features serve as nodes connected by edges encoding their spatial distance. Each graph has a matrix $\mathcal{V}$ for nodes and $\mathcal{A}$ for edges. Our task is to learn the mapping: $f_{g}$: ($\mathcal{V}_{i}$, $\mathcal{A}_{i}$) $\mapsto$ $y_{i}$. We refer to these models as \textbf{GCN}. Since we model image connections with scalars, the GCN uses a convex combination of the adjacency and identity matrix to create a filter $\mathcal{H}$ that convolves $\mathcal{V}$ before passing the output through a ReLU activation and Dropout (\textbf{Graph-Conv}). Graph Embed Pooling (\textbf{Graph-Pool}) is the corollary for MaxPooling and treats the Graph-Conv output as an embedding matrix, reducing $\mathcal{V}$ and $\mathcal{A}$ to a desired number of vertices. 

\subsubsection{Node Representations}
Each image in a cluster is represented by a node, which is composed of CNN features from ResNet-ImageWise, detected object counts ($o_i^j$ in Obj-ClusterWise), or the combination of the two. The node representations for any cluster $i$ is $\mathcal{V}_i \in \mathbb{R}^{n_i \times d}$, where $n_i$ is the maximum number of images per cluster and $d$ is the size of feature vector for each image. 
\subsubsection{Modeling Connections Between Nodes}
There is an inherent spatial structure to clusters as Mapillary images uploaded by users are geo-tagged and captured while driving or walking on roads. We take advantage of this structure by connecting the image nodes. We initialize $\mathcal{A}$ as the normalized inverse distance between two images in a cluster. That is, let $d_i^{jk}=distance(x_{i}^{j}, x_i^k)$ be the spatial distance in kilometer unit between two images $x_i^j$ and $x_i^k$ in cluster $i$. Let the maximum distance between any two images in any one cluster is $d_{max}$. In this case, for $a_i^{jk} \in \mathcal{A}_{i}$, $a_i^{jk}=1-\frac{d_i^{jk}}{d_{max}}$. This way, we construct matrix $\mathcal{A}$ using a scalar for each edge and we get: $\mathcal{A}_i \in \mathcal{R}^{n_i}$ for any cluster $i$.

%% file: sections/experiments.tex
\begin{table*}[!h] 
\begin{center}
    \resizebox{0.88\linewidth}{!}{
    \begin{tabular}{@{}ccccccccc@{}}
        \multicolumn{3}{c}{\textbf{}} & \multicolumn{2}{c}{\textbf{Wealth}} & \multicolumn{2}{c}{\textbf{Population}} & \multicolumn{2}{c}{\textbf{BMI}} \\ 
        \toprule
        \textbf{Eval Region} & \textbf{Feature} & \textbf{Method} & \textbf{Acc.} & \textbf{$r^2$} & \textbf{Acc.} & \textbf{$r^2$} & \textbf{Acc.} & \textbf{$r^2$} \\ 
        \toprule
        India & Image & ResNet-ImageWise~\cite{Suel2019london} &     74.34 & 0.51 & 93.50 & 0.85 & 85.28 & 0.52 \\
        India & Object Counts & MLP (3-layer) & 74.98 & 0.52 & 91.71 & 0.81 & 82.60 & 0.53 \\
        India & Object Counts & Random Forest & 75.77 & 0.52 & 88.99 & 0.79 & 83.49 & 0.54 \\
        India & Object Counts & GBDT & 74.91 & 0.51 & 89.35 & 0.78 & 83.63 & 0.52 \\
        India & Object Counts & kNN (k=3) & 68.69 & 0.34 & 85.78 & 0.75 & 77.98 & 0.36 \\
        India & Object Counts & GCN & 72.05 & 0.39 & 86.63 & 0.86 & 80.13 & 0.38 \\  
        India & CNN Embeddings & GCN & 81.06  & \textbf{0.54} & \textbf{94.71} & 0.82 & 89.42 & \textbf{0.57} \\ 
        India & CNN Embeddings + Object Counts & GCN & 80.91 & 0.53 & 94.42 & \textbf{0.89} & \textbf{89.56} & 0.56 \\
        
        \midrule
        India & Object Counts & Random Forest 50\% Pooled & 74.20 & 0.45 & 86.49 & 0.67 & \multicolumn{2}{c}{N/A} \\
        India & CNN Embeddings + Object Counts & GCN 50\% Pooled & 80.99 & 0.52 & 94.07 & 0.81 \\
        India & Object Counts & Random Forest 100\% Pooled  & 74.98 & 0.52 & 89.78 & 0.76 \\
        India & CNN Embeddings + Object Counts & GCN 100\% Pooled & \textbf{81.34} & 0.51 & 94.50 & 0.88\\        
        
        \midrule
        \midrule
        Kenya & Image & ResNet-ImageWise~\cite{Suel2019london} &     73.71 & 0.39 & 87.79 & \textbf{0.90}  & \multicolumn{2}{c}{N/A} \\
        Kenya & Object Counts & MLP (3-layer) & \textbf{77.34} & 0.43 & 89.59 & 0.81 \\
        Kenya & Object Counts & Random Forest & 75.59 & \textbf{0.50} & 86.38 & 0.81 \\
        Kenya & Object Counts & GBDT & 72.77 & \textbf{0.50} & 86.38 & 0.81 \\
        Kenya & Object Counts & kNN (k=3) & 70.42 & 0.38 & 82.63 & 0.53 \\
        Kenya & Object Counts & GCN & 71.36 & 0.35 & 88.26 & 0.80  \\  
        Kenya & CNN Embeddings & GCN & 76.06  & 0.42 & 89.67 & 0.84 \\ 
        Kenya & CNN Embeddings + Object Counts & GCN & 75.59 & 0.40 & 89.67 & 0.85 \\

        \midrule
        Kenya  & Object Counts & Random Forest 50\% Pooled & 74.18 & 0.41 & 81.69 & 0.51 & \multicolumn{2}{c}{N/A} \\
        Kenya & CNN Embeddings + Object Counts & GCN 50\% Pooled & 76.53 & 0.38 & 89.20 & 0.79 \\
        Kenya & Object Counts & Random Forest 100\% Pooled & 74.18 & 0.47 & 86.85 & 0.58 \\
        Kenya & CNN Embeddings + Object Counts & GCN 100\% Pooled & 77.00 & 0.36 & \textbf{90.14} & 0.72\\            
        
        \midrule
        \midrule
        Avg of Both & None (Baseline) & Random & 50.70 & - & 50.11 & - & 51.47 & -  \\
        Avg of Both & Lat/Lon Coords (Baseline) & Avg of Neighbors & 63.57 & 0.16 & 69.06 & 0.66 & 66.17 & 0.25
        \\ \bottomrule
        \end{tabular}
    }
\end{center}
\caption{
Results on predicting wealth, population, and BMI. 
The method in \cite{Suel2019london} serves as an additional baseline. To the best of our knowledge, we are the first to present a scalable, intereptable pipeline to make predictions from only crowdsourced street-level imagery across India and Kenya.}
\label{tab:classification_results}
\end{table*}

We perform extensive experiments on our dataset consisting of Mapillary images and ground-truth indexes. As our work is the first to utilize Mapillary for indicator prediction, we build baselines to benchmark our model performance. The metrics are classification accuracy and the square of the Pearson correlation coefficient (Pearson's $r^2$) for regression. For each country, we randomly sample 80\% of the clusters as the training set and the remainder is the validation set.

\subsection{Baselines}
For each cluster, we predict a local (geographic) average of neighboring clusters as a baseline. This simulates how we often have access to district-level statistics about livelihood indicators. We approximate the district-level values as the mode or mean (in the case of binary classification and regression, respectively) of the 1,000 clusters closest to $c_i$. In other words, the baseline predicts the $\hat{y}$ from $\mathcal{Y}_{i}^{sort} \subseteq \mathcal{Y}$, where $\mathcal{Y}_{i}^{sort}$ represents a subset of clusters sorted by distance to $c_i$ in increasing order where $|\mathcal{Y}_{sort}| = 1000$.


\subsection{Implementation Details}
For \textbf{ResNet-ImageWise}, all images $\mathcal{X}$ are resized to $224\times224\times3$. We train a ResNet34~\cite{he2015deep} model initialized with weights pretrained on ImageNet~\cite{russakovsky2015imagenet} to learn a mapping $f_r$ from image to $y_i \in \mathbb{R}^K$. There is one classification and one regression label for each indicator ($K = 6$). We train with batch size 128 and learning rate 0.001 (after trying 0.1, 0.01, and 0.0001) for 50 epochs on a NVIDIA 1080TI GPU with 40G of memory.

We train different models as part of \textbf{Obj-ClusterWise}, which map the object counts feature to the label space. For the classification task, we use a 3-layer Multi-Layer Perceptron (layer size 256, ReLU activation, learning rate of 0.001), Random Forest (300 trees), Gradient Boosted Decision Trees (300 boosting stages), and k-Nearest Neighbors ($k=3$). The same models are used for regression.

For the \textbf{GCN}, we feed the representation of image nodes via $\mathcal{V}$ and node connections via $\mathcal{A}$ into two Graph-Conv layers of size 64 (each followed by a ReLU activation) followed by a Graph-Pool of size 32. Then there is another pair of Graph-Conv layers of size 32 and a Graph-Pool of size 8. The resulting output is then fed into a fully connected dense layer of size 256 and a final output layer of size 2 for binary classification or size 1 for regression. We trained with a batch size of 256 and learning rate of 0.0001 on a NVIDIA 1080TI GPU. We use Adam optimizer~\cite{kingma2014adam} for all the experiments in this study. \\
\textbf{Pooled Models for Transferability} 
We also train models that pool data from both India and Kenya to assess whether learning transfers. We train a Obj-ClusterWise model and a GCN model in two experimental settings. In the first, we randomly sample 50\% of clusters from the training datasets of both countries, train the model, and evaluate on India and Kenya separately (\textbf{50\% Pooled}). In the second setting, the models learn from all training clusters from both countries and evaluate on India and Kenya separately (\textbf{100\% Pooled}). 
\subsection{Predictions on Livelihood Indicators}
\subsubsection{Poverty} 
ResNet-ImageWise, which is the baseline method from \cite{Suel2019london}, achieves 74.34\% accuracy in India and 73.71\% in Kenya.
Models that use cluster-wise representations perform better than the baselines. Obj-ClusterWise leverages the semantic information of object counts, obtaining 75.77\% in India and 77.34\% in Kenya. The GCN makes further improvements, achieving 81.06\% in India, learning from both the visual features and spatial relationships between images. The 100\% Pooled GCN achieves the best accuracy in India. As in Figure \ref{fig:distribution}, Kenya is more left-skewed than India in its wealth distribution, providing additional examples of low-wealth clusters. For regression, wealth does not persist over large areas and can shift between clusters, so the baseline does especially poorly. Obj-ClusterWise methods achieved $r^2$ scores of 0.52 to 0.54 in India and the best score of 0.50 in Kenya, suggesting the semantic information encoded by object counts is useful. 
\subsubsection{Population}
The GCN attained 94.71\% accuracy in India and 89.67\% accuracy in Kenya. The 100\% Pooled GCN achieved the best performance in Kenya at 90.14\%. The GCN also obtained the highest $r^2$, 0.89, in India. We hypothesized there would be clear visual indicators of population captured by imagery and detections (e.g. instances of "person," infrastructure, or transportation), which may explain why the GCN that uses both image embeddings and object counts performs strongly (further explored in next section). 
\subsubsection{Women's BMI}
BMI is a health-related indicator that may not be obvious from imagery, but one that our models learn to predict effectively. 
BMI data was not available for Kenya in the 2014 DHS, so we report results in India. The GCN achieved 89.56\% classification accuracy and 0.57 $r^2$. Neither high nor low BMI is desirable, so for future experiments we plan to classify a healthy BMI, between 18.5 to 24.9 \footnote{Guide to DHS Statistics, p. 154. \url{https://dhsprogram.com/pubs/pdf/DHSG1/Guide_to_DHS_Statistics_29Oct2012_DHSG1.pdf}}.

We observe the method from \cite{Suel2019london} is not adequate and make significant improvements on most tasks by leveraging cluster-wise representations. The GCN, which is designed to learn relationships between images through feature-rich nodes and spatial distance-based edges, attains the best performance on most tasks. Obj-ClusterWise models perform comparably, relying only on object counts, suggesting they capture useful semantic information. Pooled models achieve the best or close to best performance for most tasks, so learning a combination of spatial information, object counts, and visual features can be useful to generalize across countries, which will be useful for practitioners.


\subsection{Analysis: Effect of Number of Images}
\begin{figure}[!h]
    \begin{center}
    \includegraphics[width=0.9\columnwidth]{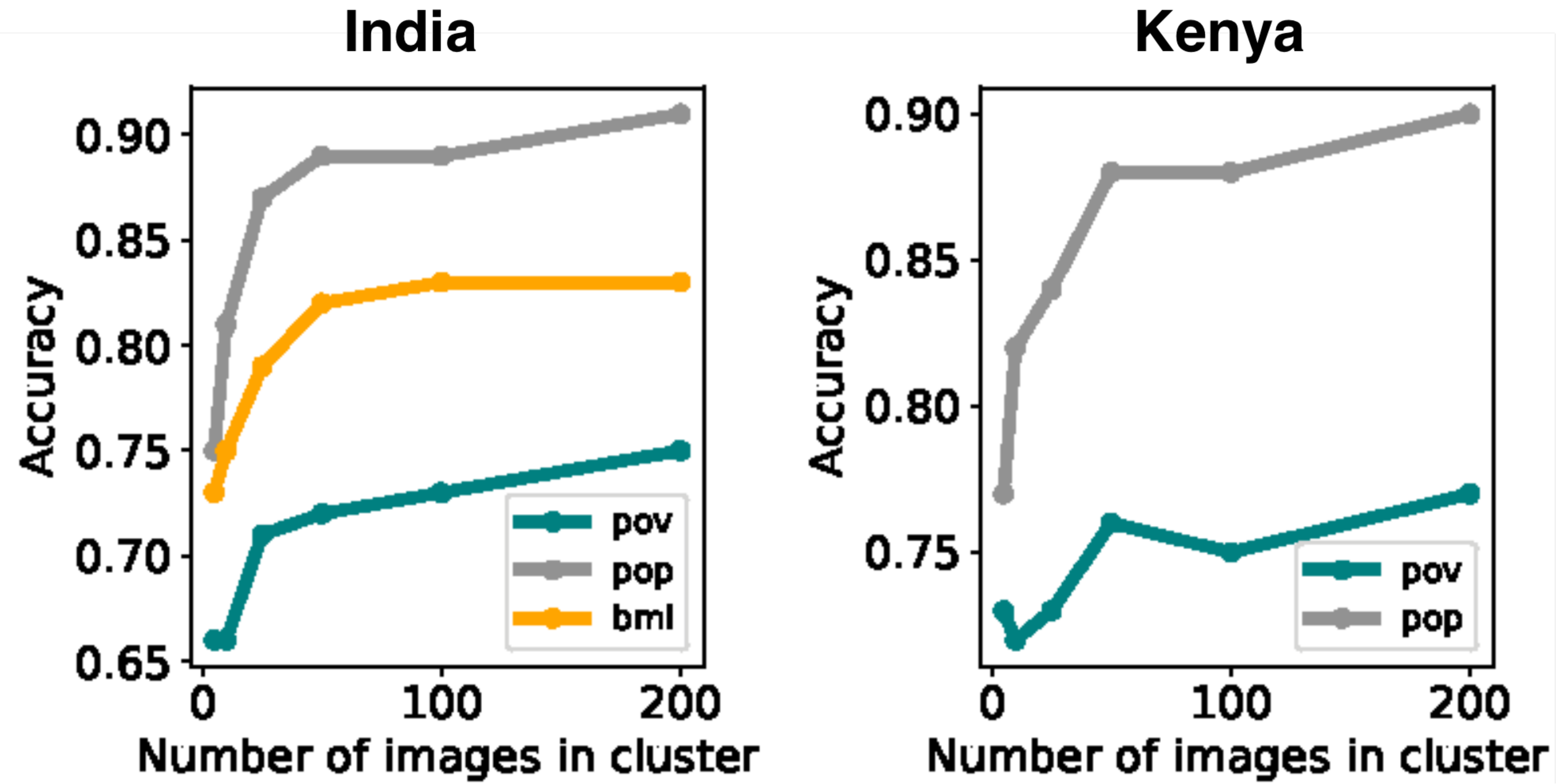}
    \end{center}
    \caption{Accuracy vs. \# of images per cluster.}
    \label{fig:clustersize}
\end{figure}

We analyzed how many images are necessary for a cluster to be sufficiently representative. We took random samples of images, from 50 to 200 (the maximum size), from each cluster. We then trained an MLP model for 100 epochs with a learning rate of 1e-3 and evaluated it on classification in each country. As shown in Figure \ref{fig:clustersize}, more images led to increased accuracy, but surprisingly a relatively small number of images, even 50, was enough to achieve good accuracy.

%% file: sections/interpretability.tex


Interpretability of the predictions is an important aspect to consider for estimates from machine learning models to be adopted by policymakers and practitioners~\cite{ayush2020generating,Murdoch22071}. 
Currently, the DHS constructs a wealth index using principal component analysis (PCA) on hand-collected characteristics of households, such as assets (i.e. televisions, bicycles, etc.), materials for housing construction, and types of water access \footnote{ \url{https://dhsprogram.com/topics/wealth-index}}. 
We demonstrate how models can learn from features visible from the road, offering a much cheaper but still effective method of index estimation.
Moreover, many existing models trained from passively collected data sources~\cite{Jean790,cadamuro2019street,sheehan2019predicting} are accurate but not inherently interpretable. We examine which objects are important for a given index and visualize decision trees to help end-user organizations understand the model. Although the GCN model performs strongly, CNN features are difficult to interpret compared to semantic objects. The 100\% Pooled Random Forest model performs closely with only object counts, so we use this model for interpretation.

\subsubsection{Feature Importance} As shown in Figure \ref{fig:mostimportantfeats}, signs of development, i.e. vehicles, traffic lights, street lights, and construction, were important in poverty prediction. Instances of terrain (e.g. dirt or exposed ground alongside a road) were also informative, likely because they suggest a lack of urbanization. For population, infrastructure, i.e. rail tracks and bridges, and transportation modes, i.e. bicyclist, motorcycle, and truck, were important. For women's BMI, the most salient features were streetlights, manholes, and billboards, indicating the presence of services and development. 
\subsubsection{Visualizing Decision Trees} We train decision tree regressors with a max depth of 3 to predict poverty and then visualize them (Figure \ref{fig:trees}). At each node, moving left means the cluster has fewer than the threshold number of objects (right means $\geq$ to the threshold). Final predictions are at the leaves, where $n$ is the number of clusters with that prediction. We demonstrate how to use the trees by tracing paths:\\
\textbf{Path 1}: As expected, a cluster with few street lights is predicted as low-wealth, shown by how most leaves have negative values. Billboards, representing storefronts and ads, are also informative, with fewer leading to a negative estimate.\\
\textbf{Path 2}: Few buildings and barrier walls leads to a low-wealth prediction. These construction-type barrier walls are salient (used in the second level of the tree), indicating sites of development and growth for organizations to monitor.\\
\textbf{Path 3}: Alternatively, many buildings but few barrier walls signal varying levels of development. Street lights become important, as more instances correlate with higher estimates.



%% file: sections/conclusion.tex
In this work, we present a novel approach to make predictions on poverty, population, and women's body-mass index from street-level imagery. In spite of the inconsistent quality of such a large-scale, crowd-sourced dataset, we achieve strong performance in predicting indicators that have not previously been the focus of machine learning methods, such as women's BMI, a key nutritional indicator. We demonstrate how our method is scalable, making predictions across India and Kenya. We present two approaches: (1) cluster-wise learning, which represents household clusters from detected object counts and enables interpretability and (2) a graph-based approach that aims to capture the spatial structure of a cluster, representing images as feature-rich nodes connected by edges based on spatial distance. We hope that our method can be employed as a cheap but effective alternative to traditional surveying methods for organizations to measure the well-being of developing regions. 
